\documentclass[11pt]{article}
\usepackage{eacl2017}
\usepackage{times}
\usepackage{color}
\usepackage{url}
\usepackage{latexsym}
\usepackage{tikz}
\usepackage{standalone}
\usepackage{amsmath}
\usepackage{multirow}

\eaclfinalcopy 

\title{Distinguishing Antonyms and Synonyms\\in a Pattern-based Neural Network}

\author{Kim Anh Nguyen \and Sabine Schulte im Walde \and Ngoc Thang Vu \\
	    Institut f\"ur Maschinelle Sprachverarbeitung\\
	    Universit\"at Stuttgart\\
	    Pfaffenwaldring 5B, 70569 Stuttgart, Germany\\
	    {\{\tt nguyenkh,schulte,thangvu\}@ims.uni-stuttgart.de}}

\date{}

\begin{document}

\maketitle

\begin{abstract}
  Distinguishing between antonyms and synonyms is a key task to
  achieve high performance in NLP systems. While they are notoriously
  difficult to distinguish by distributional co-occurrence models,
  pattern-based methods have proven effective to differentiate between
  the relations. In this paper, we present a novel neural network
  model \textit{AntSynNET} that exploits lexico-syntactic patterns
  from syntactic parse trees. In addition to the lexical and syntactic
  information, we successfully integrate the distance between the
  related words along the syntactic path as a new pattern feature. The
  results from classification experiments show that AntSynNET improves 
  the performance over prior pattern-based
  methods.
\end{abstract}

\section{Introduction}
\label{sec:intro}

Antonymy and synonymy represent lexical semantic relations that are
central to the organization of the mental
lexicon~\cite{Miller/Fellbaum:91}. While antonymy is defined as the
oppositeness between words, synonymy refers to words that are similar
in meaning~\cite{Deese:65,Lyons1977}. From a computational point of
view, distinguishing between antonymy and synonymy is important for
NLP applications such as Machine Translation and Textual Entailment,
which go beyond a general notion of semantic relatedness and require
to identify specific semantic relations. However, due to
interchangeable substitution, antonyms and synonyms often occur in
similar contexts, which makes it challenging to automatically
distinguish between them.

Two families of approaches to differentiate between antonyms and
synonyms are predominent in NLP. Both make use of distributional
vector representations, relying on the \textit{distributional
  hypothesis}~\cite{Harris1954,Firth:57}, that words with similar
distributions have related meanings: co-occurrence models and
pattern-based models. These distributional semantic models (DSMs)
offer a means to represent meaning vectors of words or word pairs, and
to determine their semantic relatedness
\cite{Turney/Pantel2010}.

In \textit{co-occurrence models}, each word is represented by a
weighted feature vector, where features typically correspond to words
that co-occur in particular contexts.  When using word embeddings,
these models rely on neural methods to represent words as
low-dimensional vectors. To create the word embeddings, the models
either make use of neural-based techniques, such as the skip-gram
model~\cite{Mikolov2013b}, or use matrix
factorization~\cite{Pennington:14} that builds word embeddings by
factorizing word-context co-occurrence matrices. In comparison to
standard co-occurrence vector representations, word embeddings address
the problematic sparsity of word vectors and have achieved impressive
results in many NLP tasks such as word similarity (e.g., \newcite{Pennington:14}),
relation classification (e.g., \newcite{ThangVu:16}), and antonym-synonym distinction (e.g., \newcite{Nguyen:16}).

In \textit{pattern-based models}, vector representations make use of
lexico-syntactic surface patterns to distinguish between the relations
of word pairs.  For example, \newcite{Justeson/Katz:91} suggested that
adjectival opposites co-occur with each other in specific linear
sequences, such as \texttt{between X and Y}. \newcite{Hearst:92}
determined surface patterns, e.g., \texttt{X such as Y}, to identify
nominal hypernyms.  \newcite{Lin2003} proposed two textual patterns
indicating semantic incompatibility, \texttt{from X to Y} and
\texttt{either X or Y}, to distinguish opposites from semantically
similar words. \newcite{Roth2014} proposed a method that combined
patterns with discourse markers for classifying paradigmatic relations
including antonymy, synonymy, and hypernymy. Recently,
\newcite{Schwartz:15} used two prominent patterns from
\newcite{Lin2003} to learn word embeddings that distinguished antonyms
from similar words in determining degrees of similarity and word
analogy.

In this paper, we present a novel pattern-based neural method
\textit{AntSynNET} to distinguish antonyms from synonyms. We
hypothesize that antonymous word pairs co-occur with each other in
lexico-syntactic patterns within a sentence more often than would be
expected by synonymous pairs.
This hypothesis is inspired by corpus-based studies on antonymy and
synonymy. Among others, \newcite{Charles1989} suggested that
adjectival opposites co-occur in patterns; \newcite{Fellbaum:95}
stated that nominal and verbal opposites co-occur in the same sentence
significantly more often than chance; \newcite{Lin2003} argued that if
two words appear in clear antonym patterns, they are unlikely to
represent synonymous pair.

We start out by inducing patterns between \texttt{X} and \texttt{Y}
from a large-scale web corpus, where \texttt{X} and \texttt{Y}
represent two words of an antonym or synonym word pair, and the
pattern is derived from the simple paths between \texttt{X} and \texttt{Y} in
a syntactic parse tree.
Each node in the simple path combines lexical and syntactic information; in
addition, we suggest a novel feature for the patterns, i.e., the
distance between the two words along the syntactic path. All pattern
features are fed into a recurrent neural network with long short-term
memory (LSTM) units~\cite{Hochreiter/Schmidhuber:97}, which encode the
patterns as vector representations. Afterwards, the vector
representations of the patterns are used in a classifier to
distinguish between antonyms and synonyms. The results from
experiments 
show that AntSynNET improves the performance over prior pattern-based methods.
Furthermore, the implementation of our models is made publicly available\footnote{\scriptsize \url{https://github.com/nguyenkh/AntSynNET}}. 

The remainder of this paper is organized as follows:
In Section~\ref{sec:related-works}, we present previous work distinguishing antonyms and synonyms.
Section~\ref{sec:AntSynNET} describes our proposed AntSynNET model.
We present the induction of the patterns
(Section~\ref{subsec:build-patterns}), describe the recurrent neural
network with long short-term memory units which is used to encode patterns within a vector representation (Section~\ref{subsec:lstm}),
and describe two models to classify antonyms and synonyms: the pure
pattern-based model (Section~\ref{subsec:pattern-model}) and the
combined model (Section~\ref{subsec:combined-model}). 
After introducing two baselines in Section~\ref{sec:baselines}, we describe
our dataset, experimental settings, results of our methods, the
effects of the newly proposed distance feature, and the effects of the
various types of word embeddings. Section~\ref{sec:conclusion} concludes the paper.

\section{Related Work}
\label{sec:related-works}

\paragraph{Pattern-based methods:} Regarding the task of
antonym-synonym distinction, there exist a variety of approaches which
rely on patterns. \newcite{Lin2003} used bilingual dependency triples
and patterns to extract distributionally similar words. They relied on
clear antonym patterns such as \texttt{from X to Y} and \texttt{either
  X or Y} in a post-processing step to distinguish antonyms from
synonyms. The main idea is that if two words \texttt{X} and \texttt{Y}
appear in one of these patterns, they are unlikely to represent
synonymous pair. \newcite{SchulteimWalde2013} proposed a method to
distinguish between the paradigmatic relations antonymy, synonymy and
hypernymy in German, based on automatically acquired word
patterns. \newcite{Roth2014} combined general lexico-syntactic
patterns with discourse markers as indicators for the same relations,
both for German and for English. They assumed that if two phrases
frequently co-occur with a specific discourse marker, then the
discourse relation expressed by the corresponding marker should also
indicate the relation between the words in the affected phrases. By
using the raw corpus and a fixed list of discourse markers, the model
can easily be extended to other languages. More recently,
\newcite{Schwartz:15} presented a symmetric pattern-based model for
word vector representation in which antonyms are assigned to
dissimilar vector representations. Differently to the previous
pattern-based methods which used the standard distribution of
patterns, Schwartz et al. used patterns to learn word embeddings.

\paragraph{Vector representation methods:} \newcite{Yih:12} introduced
a new vector representation where antonyms lie on opposite sides of a
sphere. They derived this representation with the incorporation of a
thesaurus and latent semantic analysis, by assigning signs to the
entries in the co-occurrence matrix on which latent semantic analysis
operates, such that synonyms would tend to have positive cosine
similarities, and antonyms would tend to have negative cosine
similarities. \newcite{Scheible2013} showed that the distributional
difference between antonyms and synonyms can be identified via a
simple word space model by using appropriate features. Instead of
taking into account all words in a window of a certain size for
feature extraction, the authors experimented with only words of a
certain part-of-speech, and restricted
distributions. \newcite{Santus2014b} proposed a different method to
distinguish antonyms from synonyms by identifying the most salient
dimensions of meaning in vector representations and reporting a new
average-precision-based distributional measure and an entropy-based
measure. \newcite{Ono2015} trained supervised word embeddings for the
task of identifying antonymy. They proposed two models to learn word
embeddings: the first model relied on thesaurus information; the
second model made use of distributional information and thesaurus
information.  More recently, \newcite{Nguyen:16} proposed two methods
to distinguish antonyms from synonyms: in the first method, the
authors improved the quality of weighted feature vectors by
strengthening those features that are most salient in the vectors, and
by putting less emphasis on those that are of minor importance when
distinguishing degrees of similarity between words. In the second
method, the lexical contrast information was integrated into the
skip-gram model~\cite{Mikolov2013b} to learn word embeddings. This
model successfully predicted degrees of similarity and identified
antonyms and synonyms.
\section{AntSynNET: LSTM-based Antonym-Synonym Distinction}
\label{sec:AntSynNET}
In this section, we describe the AntSynNET model, using a
pattern-based LSTM for distinguishing antonyms from synonyms. We first
present the induction of patterns from a parsed corpus
(Section~\ref{subsec:build-patterns}). Section~\ref{subsec:lstm} then
describes how we utilize the recurrent neural network with long
short-term memory units to encode the patterns as vector
representation. Finally, we present the AntSynNET model and two
approaches to classify antonyms and synonyms
(Section~\ref{subsec:AntSynNET-model}).

\subsection{Induction of Patterns}
\label{subsec:build-patterns}

\begin{figure*}[t]
	\centering
	\includegraphics[width=0.75\textwidth]{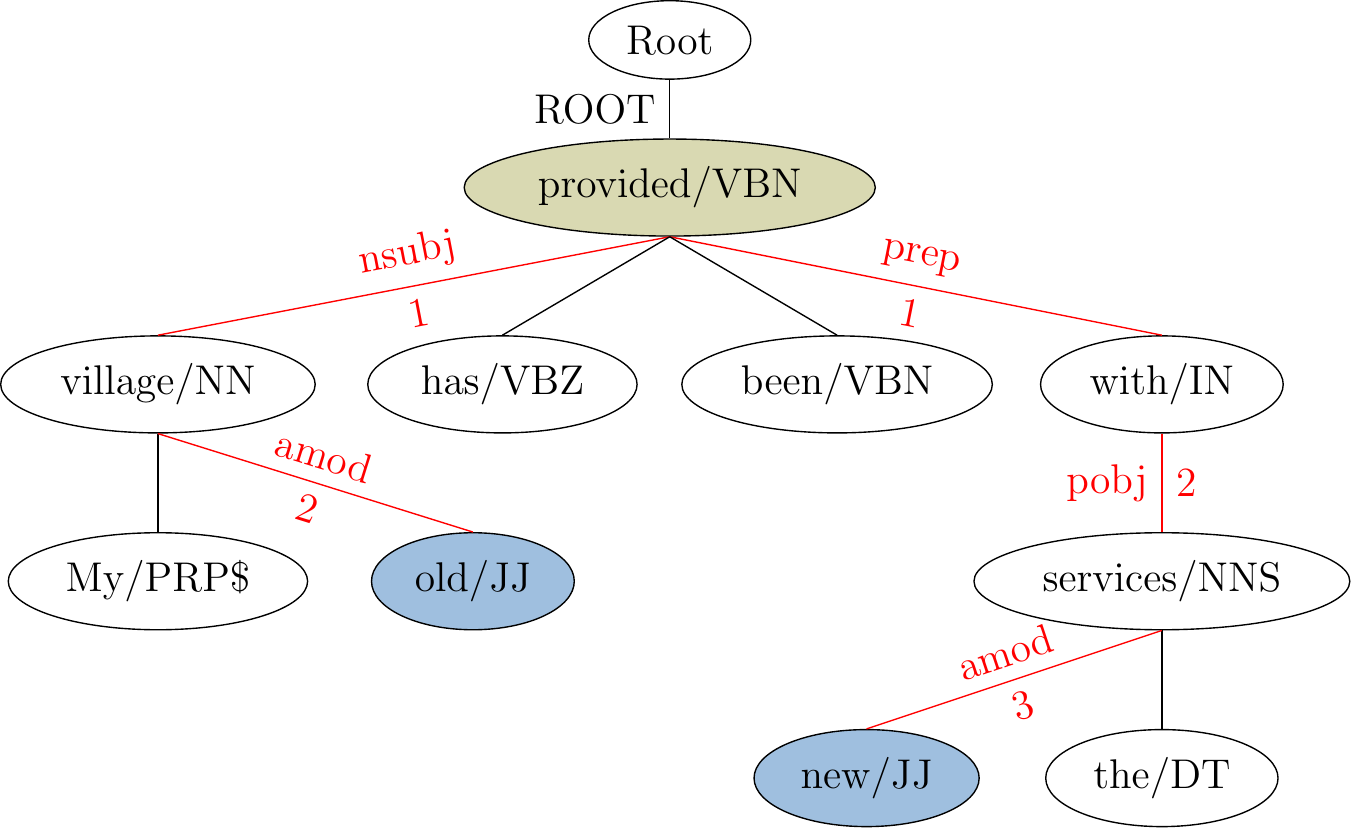} %
	\caption{Illustration of the syntactic tree for the sentence
          \textit{``My old village has been provided with the new
            services"}. Red lines indicate the path from the word
          \texttt{old} to the word \texttt{new}.}
	\label{fig:tree}
\end{figure*}
\begin{figure*}[t]
	\centering
	\includegraphics[width=.99\textwidth]{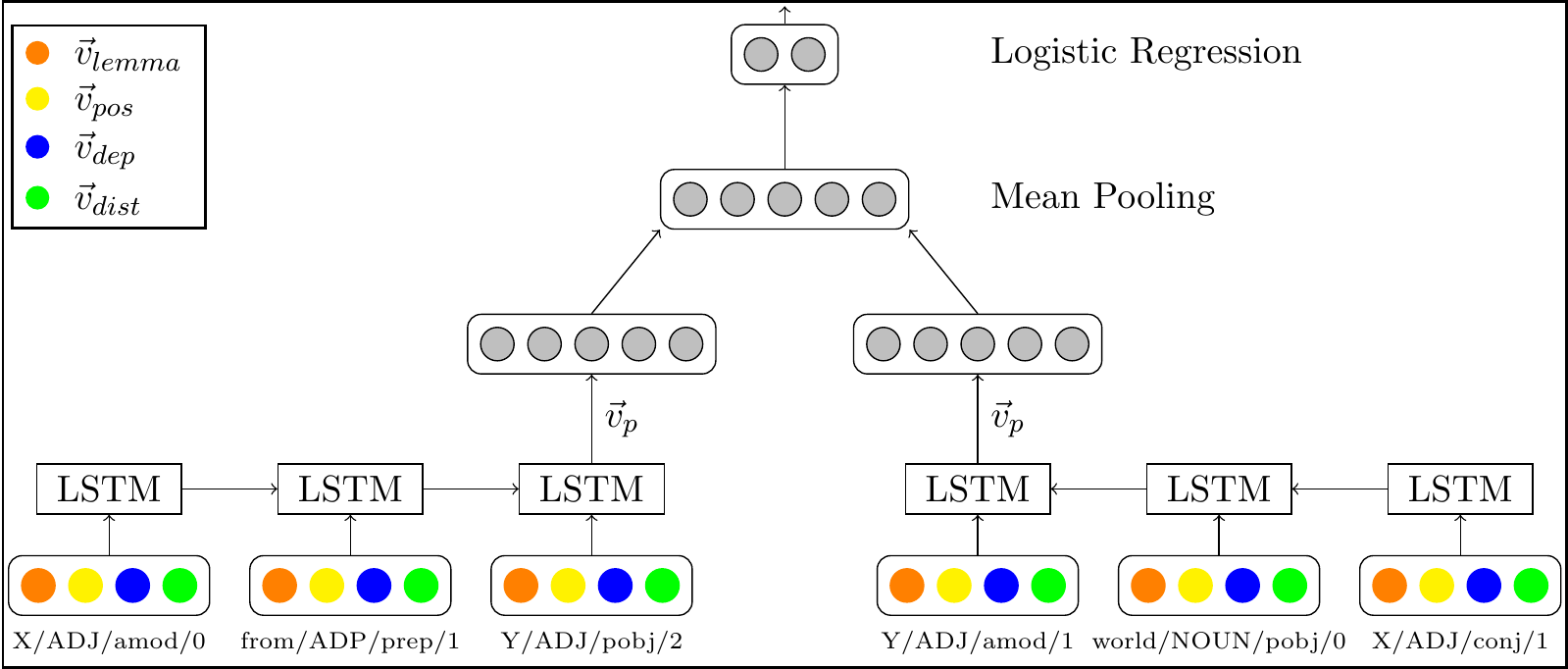} %
	\caption{Illustration of the \textit{AntSynNET} model. Each
          word pair is represented by several patterns, and each
          pattern represents a path in the graph of the syntactic
          tree. Patterns consist of several nodes where each node is
          represented by a vector with four features: lemma, POS,
          dependency label, and distance label. The mean pooling of the
          pattern vectors is the vector representation of each word
          pair, which is then fed to the logistic regression layer to
          classify antonyms and synonyms.}
	\label{fig:model}
\end{figure*}

Corpus-based studies on antonymy have suggested that opposites
co-occur with each other within a sentence significantly more often
than would be expected by chance. Our method thus makes use of
patterns as the main indicators of word pair co-occurrence, to enforce
a distinction between antonyms and synonyms. Figure~\ref{fig:tree}
shows a syntactic parse tree of the sentence \textit{``My old village
  has been provided with the new services''}. Following the
characterizations of a tree in graph theory, any two nodes (vertices)
of a tree are connected by a simple path (or one unique path). The
simple path is the shortest path between any two nodes in a tree and
does not contain repeated nodes.  In the example, the lexico-syntactic
tree pattern of the antonymous pair \textit{old--new} is determined by
finding the simple path (in red) from the lemma \texttt{old} to the
lemma \texttt{new}.  It focuses on the most relevant information and
ignores irrelevant information which does not appear in the simple
path (i.e., \textit{has, been}).
The example pattern between \texttt{X = old} and \texttt{Y = new} in
Figure~\ref{fig:tree} is represented as follows: {\small
  \texttt{X/JJ/amod/2 -- village/NN/nsubj/1 -- provide/VBN/ROOT/0 --
    with/IN/prep/1 -- service/NNS/pobj/2 -- Y/JJ/amod/3}.}

\paragraph{Node Representation:} The path patterns make use of four
features to represent each node in the syntax tree: lemma,
part-of-speech (POS) tag, dependency label and distance label. The
lemma feature captures the lexical information of words in the
sentence, while the POS and dependency features capture the
morpho-syntactic information of the sentence. The distance label
measures the path distance between the target word nodes in the
syntactic tree. Each step between a parent and a child node represents
a distance of 1; and the ancestor nodes of the remaining nodes in the
path are represented by a distance of 0. For example, the node
\texttt{provided} is an ancestor node of the simple path from
\texttt{old} to \texttt{new}. The distances from the node
\texttt{provided} to the nodes \texttt{village} and \texttt{old} are 1
and 2, respectively.
%

The vector representation of each node concatenates the four-feature
vectors as follows:
\[
  \vec{v}_{node} = [\vec{v}_{lemma} \oplus \vec{v}_{pos} \oplus \vec{v}_{dep} \oplus \vec{v}_{dist}]	
\]
where $\vec{v}_{lemma}, \vec{v}_{pos}, \vec{v}_{dep}, \vec{v}_{dist}$
represent the embeddings of the lemma, POS tag, dependency label and
distance label, respectively; and the $\oplus$ denotes the
concatenation operation.

\paragraph{Pattern Representation:} For a pattern $p$ which is
constructed by the sequence of nodes $n_1, n_2, ..., n_k$, the pattern
representation of $p$ is a sequence of vectors:
$p=[\vec{n}_1, \vec{n}_2, ..., \vec{n}_k]$. The pattern vector
$\vec{v}_p$ is then encoded by applying a recurrent neural network.

\subsection{Recurrent Neural Network with Long Short-Term Memory Units}
\label{subsec:lstm}

A recurrent neural network (RNN) is suitable for modeling sequential
data by a vector representation. In our methods, we use a long
short-term memory (LSTM) network, a variant of a recurrent neural
network to encode patterns, for the following reasons. Given a
sequence of words $p=[n_1,n_2,...,n_k]$ as input data, an RNN
processes each word $n_t$ at a time, and returns a vector of state
$h_k$ for the complete input sequence. For each time step $t$, the RNN
updates an internal memory state $h_t$ which depends on the current
input $n_t$ and the previous state $h_{t-1}$. Yet, if the sequential
input is a long-term dependency, an RNN faces the problem of gradient
vanishing or exploding, leading to difficulties in training the model.

LSTM units address these problems. The underlying idea of an LSTM is
to use an adaptive gating mechanism to decide on the degree that
LSTM units keep the previous state and memorize the extracted features
of the current input. More specifically, an LSTM comprises four
components: an input gate $i_t$, a forget gate $f_t$, an output gate
$o_t$, and a memory cell $c_t$. The state of an LSTM at each time step
$t$ is formalized as follows:
\[
  \begin{array}{l}
    i_t = \sigma(W_i \cdot x_t + U_i \cdot h_{t-1} + b_i) \\
    f_t = \sigma(W_f \cdot x_t + U_f \cdot h_{t-1} + b_f) \\ 
    o_t = \sigma(W_o \cdot x_t + U_o \cdot h_{t-1} + b_o) \\
    g_t = \tanh(W_c \cdot x_t + U_c \cdot h_{t-1} + b_c) \\
    c_t = i_t \otimes g_t + f_t \otimes c_{t-1}
  \end{array}
\]
$W$ refers to a matrix of weights that projects information between
two layers; $b$ is a layer-specific vector of bias terms; $\sigma$
denotes the sigmoid function. The output of an LSTM at a time step $t$
is computed as follows:
\[
  h_t = o_t \otimes \tanh(c_t)
\]
where $\otimes$ denotes element-wise multiplication. In our methods,
we rely on the last state $h_k$ to represent the vector $\vec{v}_p$ of
a pattern $p = [\vec{n}_1, \vec{n}_2,...,\vec{n}_k]$.

\subsection{The Proposed AntSynNET Model}
\label{subsec:AntSynNET-model}

In this section, we present two models to distinguish antonyms from
synonyms. The first model makes use of patterns to classify antonyms
and synonyms, by using an LSTM to encode patterns as vector
representations and then feeding those vectors to a logistic
regression layer (Section~\ref{subsec:pattern-model}). The second
model creates combined vector representations of word pairs, which
concatenate the vectors of the words and the patterns
(Section~\ref{subsec:combined-model}).

\subsubsection{Pattern-based AntSynNET}
\label{subsec:pattern-model}

In this model, we make use of a recurrent neural network with LSTM
units to encode patterns containing a sequence of
nodes. Figure~\ref{fig:model} illustrates the AntSynNET model. Given a
word pair $(x,y)$, we induce patterns for $(x,y)$ from a corpus, where
each pattern represents a path from $x$ to $y$
(cf. Section~\ref{subsec:build-patterns}). We then feed each pattern
$p$ of the word pair $(x,y)$ into an LSTM to obtain $\vec{v}_p$, the
vector representation of the pattern $p$
(cf. Section~\ref{subsec:lstm}). For each word pair $(x,y)$, the
vector representation of $(x,y)$ is computed as follows:
\begin{equation}
  \vec{v}_{xy} = \frac{\sum \nolimits_{p \in P(x,y)} {\vec{v}_p \cdot c_p}}
  {\sum \nolimits_{p \in P(x,y)} {c_p}}
  \label{eq:pattern}
\end{equation}
$\vec{v}_{xy}$ refers to the vector of the word pair $(x,y)$; $P(x,y)$
is the set of patterns corresponding to the pair $(x,y)$; $c_p$ is the
frequency of the pattern $p$. The vector $\vec{v}_{xy}$ is then fed
into a logistic regression layer whose target is the class label
associated with the pair $(x,y)$. Finally, the pair $(x,y)$ is
predicted as positive (i.e., antonymous) word pair if the probability
of the prediction for $\vec{v}_{xy}$ is larger than 0.5.

\subsubsection{Combined AntSynNET}
\label{subsec:combined-model}

Inspired by the supervised distributional concatenation method
in~\newcite{Baroni:12} and the integrated path-based and distributional
method for hypernymy detection in~\newcite{Shwartz:16}, we take into
account the patterns and distribution of target pairs to create their
combined vector representations. Given a word pair $(x,y)$, the
combined vector representation of the pair $(x,y)$ is determined by
using both the co-occurrence distribution of the words and the
syntactic path patterns:
\begin{equation}
  \vec{v}_{comb(x,y)} = [\vec{v}_x \oplus \vec{v}_{xy} \oplus \vec{v}_y]
  \label{eq:combined}
\end{equation}
$\vec{v}_{comb(x,y)}$ refers to the combined vector of the word pair
$(x,y)$; $\vec{v}_x$ and $\vec{v}_y$ are the vectors of word $x$ and
word $y$, respectively; $\vec{v}_{xy}$ is the vector of the pattern
that corresponds to the pair $(x,y)$,
cf. Section~\ref{subsec:pattern-model}. Similar to the pattern-based
model, the combined vector $\vec{v}_{comb(x,y)}$ is fed into the
logistic regression layer to classify antonyms and synonyms.

\section{Baseline Models}
\label{sec:baselines}

To compare AntSynNET with baseline models for pattern-based
classification of antonyms and synonyms, we introduce two
pattern-based baseline methods: the distributional method
(Section~\ref{subsec:baseline1}), and the distributed method
(Section~\ref{subsec:baseline2}).

\subsection{Distributional Baseline}
\label{subsec:baseline1}

As a first baseline, we apply the approach by~\newcite{Roth2014},
henceforth R\&SiW. They used a vector space model to represent pairs
of words by a combination of standard lexico-syntactic patterns and
discourse markers. In addition to the patterns, the discourse markers
added information to express discourse relations, which in turn may
indicate the specific semantic relation between the two words in a
word pair. For example, contrast relations might indicate antonymy,
whereas elaborations may indicate synonymy or hyponymy.

Michael Roth, the first author of R\&SiW, kindly computed the relation
classification results of the pattern--discourse model for our test
sets. The weights between marker-based and pattern-based models were
tuned on the validation sets; other hyperparameters were set exactly
as described by the R\&SiW method.

\subsection{Distributed Baseline}
\label{subsec:baseline2}

The \textit{SP} method proposed by \newcite{Schwartz:15} uses
symmetric patterns for generating word embeddings. In this work, the
authors applied an unsupervised algorithm for the automatic extraction
of symmetric patterns from plain text. The symmetric patterns were
defined as a sequence of 3-5 tokens consisting of exactly two
wildcards and 1-3 words. The patterns were filtered based on their
frequencies, such that the resulting pattern set contained 11
patterns. For generating word embeddings, a matrix of co-occurrence
counts between patterns and words in the vocabulary was computed,
using positive point-wise mutual information. The sparsity problem of
vector representations was addressed by smoothing. For antonym
representation, the authors relied on two patterns suggested
by~\newcite{Lin2003} to construct word embeddings containing an
antonym parameter that can be turned on in order to represent antonyms
as dissimilar, and that can be turned off to represent antonyms as
similar.

To apply the SP method to our data, we make use of the pre-trained SP
embeddings\footnote{\scriptsize \url{http://homes.cs.washington.edu/~roysch/papers/sp_embeddings/sp_embeddings.html}}
with 500 dimensions\footnote{The 500-dimensional embeddings
  outperformed the 300-dimensional embeddings for our data.}. We
calculate the cosine similarity of word pairs and then use a Support
Vector Machine with Radial Basis Function kernel to classify antonyms
and synonyms.

\section{Experiments}
\label{sec:experiments}
\subsection{Dataset}

For training the models, neural networks require a large amount of
training data. We use the existing large-scale antonym and
synonym pairs previously used by~\newcite{Nguyen:16}. Originally, the
data pairs were collected from WordNet~\cite{Miller1995} and
Wordnik\footnote{\scriptsize \url{http://www.wordnik.com}}.

In order to induce patterns for the word pairs in the dataset, we
identify the sentences in the corpus that contain the word
pair. Thereafter, we extract all patterns for the word pair. We
filter out all patterns which occur less than five times; and we only
take into account word pairs that contain at least five patterns for
training, validating and testing. For the proportion of positive and
negative pairs, we keep a ratio of 1:1 positive (antonym) to negative
(synonym) pairs in the dataset. In order to create the sets of
training, testing and validation data, we perform random splitting
with 70\% train, 25\% test, and 5\% validation sets. The final dataset
contains the number of word pairs according to word classes described in Table~\ref{dataset}. Moreover, Table~\ref{statistic} shows the average number of patterns for each word pair in our dataset.
\begin{table}[h]
  \centering
  \resizebox{0.95\columnwidth}{!}{%
    \begin{tabular}{lcccc}
      \hline
      \multicolumn{1}{c}{\textbf{Word Class}} & \textbf{Train} & \textbf{Test} & \textbf{Validation} & \textbf{Total} \\ \hline
      Adjective                               & 5562           & 1986          & 398                 & 7946           \\
      Verb                                    & 2534           & 908           & 182                 & 3624           \\
      Noun                                    & 2836           & 1020          & 206                 & 4062           \\ \hline
    \end{tabular}
  }
  \caption{Our dataset.}
  \label{dataset}
\end{table}
\begin{table}[h]
  \centering
  \resizebox{0.95\columnwidth}{!}{%
    \begin{tabular}{lccc}
      \hline
      \multicolumn{1}{c}{\textbf{Word Class}} & \textbf{Train} & \textbf{Test} & \textbf{Validation} \\ \hline
      Adjective                               & 135            & 131           & 141                 \\
      Verb                                    & 364            & 332           & 396                 \\
      Noun                                    & 110            & 132           & 105                 \\ \hline
    \end{tabular}
  }
  \caption{Average number of patterns per word pair across word classes.}
  \label{statistic}
\end{table}
\subsection{Experimental Settings}

We use the English Wikipedia
dump\footnote{\scriptsize \url{https://dumps.wikimedia.org/enwiki/latest/enwiki-latest-pages-articles.xml.bz2}}
from June 2016 as the corpus resource for our methods and
baselines. For parsing the corpus, we rely on
spaCy\footnote{\scriptsize \url{https://spacy.io}}. For the lemma embeddings, we
rely on the word embeddings of the dLCE
model\footnote{\scriptsize \url{https://github.com/nguyenkh/AntSynDistinction}}~\cite{Nguyen:16}
which is the state-of-the-art vector representation for distinguishing
antonyms from synonyms. We re-implemented this cutting-edge model
on Wikipedia with 100 dimensions, and then make use of the dLCE word
embeddings for initialization the lemma embeddings. The embeddings of
POS tags, dependency labels, distance labels, and out-of-vocabulary
lemmas are initialized randomly. The number of dimensions is set to 10
for the embeddings of POS tags, dependency labels and distance
labels. We use the validation sets to tune the number of dimensions
for these labels. For optimization, we rely on the cross-entropy loss
function and Stochastic Gradient Descent with the Adadelta update
rule~\cite{Zeiler:12}. For training, we use the \texttt{Theano}
framework~\cite{Theano}. Regularization is applied by a dropout of 0.5
on each of component's embeddings (dropout rate is tuned on the
validation set). We train the models with 40 epochs and update all
embeddings during training.
\begin{table*}[t]
\centering
\resizebox{0.95\textwidth}{!}{%
\begin{tabular}{l|ccl|ccc|ccl}
\hline
\multicolumn{1}{c|}{\multirow{2}{*}{\textbf{Model}}} & \multicolumn{3}{c|}{\textbf{Adjective}}          & \multicolumn{3}{c|}{\textbf{Verb}}               & \multicolumn{3}{c}{\textbf{Noun}}               \\ \cline{2-10} 
\multicolumn{1}{c|}{}                                & \textbf{P}     & \textbf{R}     & \multicolumn{1}{c|}{$\mathbf{F_1}$} & \textbf{P}     & \textbf{R}     & $\mathbf{F_1}$ & \textbf{P}     & \textbf{R}     & \multicolumn{1}{c}{$\mathbf{F_1}$} \\ \hline
SP baseline                                                 & 0.730          & 0.706          & 0.718          & 0.560          & 0.609          & 0.584          & 0.625          & 0.393          & 0.482          \\
R\&SiW baseline                                             & 0.717          & 0.717          & 0.717          & \textbf{0.789} & 0.787          & \textbf{0.788} & \textbf{0.833} & 0.831          & 0.832          \\
Pattern-based AntSynNET                              & \textbf{0.764} & 0.788          & 0.776$^*$          & 0.741          & \textbf{0.833} & 0.784          & 0.804          & 0.851          & 0.827          \\
Combined AntSynNET                                   & 0.763          & \textbf{0.807} & \textbf{0.784}$^*$ & 0.743          & 0.815          & 0.777          & 0.816          & \textbf{0.898} & \textbf{0.855}$^{**}$ \\ \hline
\end{tabular}
}
\caption{Performance of the AntSynNET models in comparison to the baseline models.}
\label{tbl:results}
\end{table*}
\begin{table*}[t]
\centering
\resizebox{0.95\textwidth}{!}{%
\begin{tabular}{c|l|lll|lll|lll}
\hline
\multirow{2}{*}{\textbf{Feature}} & \multicolumn{1}{c|}{\multirow{2}{*}{\textbf{Model}}} & \multicolumn{3}{c|}{\textbf{Adjective}}                                                               & \multicolumn{3}{c|}{\textbf{Verb}}                                                                    & \multicolumn{3}{c}{\textbf{Noun}}                                                                   \\ \cline{3-11} 
                                  & \multicolumn{1}{c|}{}                                & \multicolumn{1}{c}{\textbf{P}} & \multicolumn{1}{c}{\textbf{R}} & \multicolumn{1}{c|}{$\mathbf{F_1}$} & \multicolumn{1}{c}{\textbf{P}} & \multicolumn{1}{c}{\textbf{R}} & \multicolumn{1}{c|}{$\mathbf{F_1}$} & \multicolumn{1}{c}{\textbf{P}} & \multicolumn{1}{c}{\textbf{R}} & \multicolumn{1}{c}{$\mathbf{F_1}$} \\ \hline
\multirow{2}{*}{Direction}        & Pattern-based                                           & 0.752                          & 0.755                          & 0.753                               & 0.734                          & 0.819                          & 0.774                               & 0.800                          & 0.825                          & 0.813                              \\
                                  & Combined                                           & 0.754                          & 0.784                          & 0.769                               & 0.739                          & 0.793                          & 0.765                               & \textbf{0.829}                 & 0.810                          & 0.819                              \\ \hline
\multirow{2}{*}{Distance}         & Pattern-based                                        & \textbf{0.764}                 & 0.788                          & 0.776                               & 0.741                          & \textbf{0.833}                 & \textbf{0.784}                      & 0.804                          & 0.851                          & 0.827                              \\
                                  & Combined                                             & 0.763                          & \textbf{0.807}                 & \textbf{0.784}$^{**}$                     & \textbf{0.743}                 & 0.815                          & 0.777                               & 0.816                          & \textbf{0.898}                 & \textbf{0.855}$^{**}$                     \\ \hline
\end{tabular}
}
\caption{Comparing the novel distance feature with Schwarz et al.'s direction feature, across word classes.}
\label{tbl:featurel}
\end{table*}

\subsection{Overall Results}
\label{subsec:results}

Table~\ref{tbl:results} shows the significant\footnote{t-test, *$p < 0.05$, **$p < 0.1$} performance of our models in
comparison to the baselines. Concerning adjectives, the two proposed
models significantly outperform the two baselines: The performance of
the baselines is around .72 for $F_1$, and the corresponding results
for the combined AntSynNET model achieve an improvement of
$>$.06. Regarding nouns, the improvement of the new methods is just
.02 \texttt{$F_1$} in comparison to the R\&SiW baseline, but we
achieve a much better performance in comparison to the SP baseline, an
increase of .37 \texttt{$F_1$}.
Regarding verbs, we do not outperform the more advanced R\&SiW
baseline in terms of the \texttt{$F_1$} score, but we obtain higher
recall scores. In comparison to the SP baseline, our models still show
a clear \texttt{$F_1$} improvement.

Overall, our proposed models achieve comparatively high recall scores
compared to the two baselines.  This strengthens our hypothesis that
there is a higher possibility for the co-occurrence of antonymous
pairs in patterns over synonymous pairs within a sentence. Because,
when the proposed models obtain high recall scores, the models are
able to retrieve most relevant information (antonymous pairs)
corresponding to the patterns. 
Regarding the low precision in the two proposed models, 
we sampled randomly 5 pairs in each population: true positive, true negative, false positive, false negative. We then compared the overlap of patterns for the true predictions (true positive pairs and true negative pairs) and the false predictions (false positive pairs and false negative pairs). We found out that there is no overlap between patterns of true predictions; and the number overlap between patterns of false predictions is 2, 2, and 4 patterns for noun, adjective, and verb classes, respectively. This shows that the low precision of our models stems from the patterns which represent both antonymous and synonymous pairs.
%
\begin{table*}[t]
\centering
\resizebox{0.98\textwidth}{!}{%
\begin{tabular}{l|l|ccc|ccc|ccc}
\hline
\multicolumn{1}{c|}{\multirow{2}{*}{\textbf{Model}}} & \multicolumn{1}{c|}{\multirow{2}{*}{\textbf{Word Embeddings}}} & \multicolumn{3}{c|}{\textbf{Adjective}}           & \multicolumn{3}{c|}{\textbf{Verb}}                & \multicolumn{3}{c}{\textbf{Noun}}                \\ \cline{3-11} 
\multicolumn{1}{c|}{}                                & \multicolumn{1}{c|}{}                                          & \textbf{P} & \textbf{R} & \textbf{$\mathbf{F_1}$} & \textbf{P} & \textbf{R} & \textbf{$\mathbf{F_1}$} & \textbf{P} & \textbf{R} & \textbf{$\mathbf{F_1}$} \\ \hline
\multirow{2}{*}{Pattern-based Model}                 & GloVe                                                          & 0.763      & 0.770      & 0.767                   & 0.705      & 0.852      & 0.772                   & 0.789      & 0.849      & 0.818                   \\
                                                     & dLCE                                                           & 0.764      & 0.788      & \textbf{0.776}          & 0.741      & 0.833      & \textbf{0.784}          & 0.804      & 0.851      & \textbf{0.827}          \\ \hline
\multirow{2}{*}{Combined Model}                      & Glove                                                          & 0.750      & 0.798      & 0.773                   & 0.717      & 0.826      & 0.768                   & 0.807      & 0.827      & 0.817                   \\
                                                     & dLCE                                                           & 0.763      & 0.807      & \textbf{0.784}          & 0.743      & 0.815      & \textbf{0.777}          & 0.816      & 0.898      & \textbf{0.855}          \\ \hline
\end{tabular}
}
\caption{Comparing pre-trained GloVe and dLCE word embeddings.}
\label{tbl:embeddings}
\end{table*}
\subsection{Effect of the Distance Feature}

In our models, the novel distance feature is successfully integrated along the syntactic path
to represent lexico-syntactic patterns. The intuition behind the distance feature exploits properties of trees in graph theory, which show that there exist
differences in the degree of relationship between the parent node and
the child nodes ($distance = 1$) and in the degree of relationship between
the ancestor node and the descendant nodes ($distance > 1$). Hence, we use
the distance feature to effectively capture these relationships.

In order to evaluate the effect of our novel distance feature, we
compare the distance feature to the direction feature proposed
by~\newcite{Shwartz:16}. In their approach, the authors combined
lemma, POS, dependency, and direction features for the task of
hypernym detection. The direction feature represented the direction of
the dependency label between two nodes in a path from \texttt{X} to
\texttt{Y}.

For evaluation, we make use of the same information regarding dataset
and patterns as in Section~\ref{subsec:results}, and then replace the
distance feature by the direction feature. The results are shown in
Table~\ref{tbl:featurel}. The distance feature enhances the
performance of our proposed models more effectively than the direction
feature does, across all word classes.
%

\subsection{Effect of Word Embeddings}
\label{subsec:effects-word-embeddings}


Our methods rely on the word embeddings of the dLCE model,
state-of-the-art word embeddings for antonym-synonym distinction. Yet,
the word embeddings of the dLCE model, i.e., supervised word
embeddings, represent information collected from lexical resources. In
order to evaluate the effect of these word embeddings on the
performance of our models, we replace them by the pre-trained GloVe
word
embeddings\footnote{\scriptsize \url{http://www-nlp.stanford.edu/projects/glove/}}
with 100 dimensions, and compare the effects of the GloVe word
embeddings and the dLCE word embeddings on the performance of the two
proposed models.

Table~\ref{tbl:embeddings} illustrates the performance of our two
models on all word classes. The table shows that the dLCE word
embeddings are better than the pre-trained GloVe word embeddings, by
around .01 $F_1$ for the pattern-based AntSynNET model and the
combined AntSynNET model regarding adjective and verb pairs. Regarding
noun pairs, the improvements of the dLCE word embeddings over
pre-trained GloVe word embeddings achieve around .01 and .04 $F_1$ for
the pattern-based model and the combined model, respectively.

\section{Conclusion}
\label{sec:conclusion}

In this paper, we presented a novel pattern-based neural method
\textit{AntSynNET} to distinguish antonyms from synonyms. We
hypothesized that antonymous word pairs co-occur with each other in
lexico-syntactic patterns within a sentence more often than synonymous
word pairs.

The patterns were derived from the simple paths between semantically
related words in a syntactic parse tree. In addition to lexical and
syntactic information, we suggested a novel path distance feature. The
AntSynNET model consists of two approaches to classify antonyms and
synonyms.  In the first approach, we used a recurrent neural network
with long short-term memory units to encode the patterns as vector
representations; in the second approach, we made use of the
distribution and encoded patterns of the target pairs to generate
combined vector representations. The resulting vectors of patterns in
both approaches were fed into the logistic regression layer for
classification.

Our proposed models significantly outperformed two baselines relying
on previous work, mainly in terms of recall. Moreover, we demonstrated
that the distance feature outperformed a previously suggested
direction feature, and that our embeddings outperformed the
state-of-the-art GloVe embeddings. Last but not least, our two
proposed models only rely on corpus data, such that the models are
easily applicable to other languages and relations.

\section*{Acknowledgements}
We would like to thank Michael Roth for helping us to compute the results of the R\&SiW model on our dataset. 

The research was supported by the Ministry of Education and Training of the Socialist Republic of Vietnam (Scholarship 977/QD-BGDDT; Kim-Anh Nguyen), the DFG Collaborative Research Centre SFB 732 (Kim-Anh Nguyen, Ngoc Thang Vu), and the DFG Heisenberg Fellowship SCHU-2580/1 (Sabine Schulte im Walde).

\bibliography{eacl2017}
\bibliographystyle{eacl2017}

\end{document}